# Particle Swarm Optimized Fuzzy Controller for Indirect Vector Control of Multilevel Inverter Fed Induction Motor

**Sanjaya Kumar Sahu[1],T. V. Dixit [2] and D.D. Neema[3]**

**[1]Electrical Engineering, Bhilai Institute of Technology
Durg, Chhattisgarh491001, India**

**[2]Electrical Engineering, Sarguja University
Ambikapur, Chhattisgarh-497001, India**

**[3]Electrical and Electronics Engineering, Chhattisgarh Institute of Technology
Rajanandgaon, Chhattisgarh-491445, India**

**Abstract**

The Particle Swarm Optimized (PSO) fuzzy controller has been proposed for indirect vector control of induction motor. In this proposed scheme a Neutral Point Clamped (NPC) multilevel inverter is used and hysteresis current control technique has been adopted for switching the IGBTs. A Mamdani type fuzzy controller is used in place of conventional PI controller. To ensure better performance of fuzzy controller all parameters such as membership functions, normalizing and de-normalizing parameters are optimized using PSO. The performance of proposed controller is investigated under various load and speed conditions. The simulation results show its stability and robustness for high performance derives applications.

***Keywords:*** *Multilevel Inverter, Hysteresis Current Control, Particle Swarm Optimization (PSO), Fuzzy Logic Controller (FLC).*

## 1. Introduction

Three phase induction motors are widely used in the industrial purpose because they show better performance during heavy loads as well as cost effective. However the drawbacks associated with induction motor are its non-linear behaviour, controllability and its complexity in developing mathematical model [1]. By vector control or field oriented control (FOC) theory, induction motor can be controlled like a separately excited dc motor. As a result field and torque of the induction machine can be controlled independently by manipulating the corresponding field oriented quantities. There are two methods of vector control - direct and indirect vector control [1]–[3]. In this paper the indirect control method is adopted, where the slip angle, the d-axis and q-axis stator currents in synchronous reference frame are computed from the torque and rotor flux and used for vector control.

A multi-level inverter is a power electronic circuit built to synthesize stepped approximation of a sinusoidal wave output voltage or current from a number of DC voltages [4]. The multilevel inverters gained the attention in industrial drive application due to following features [5]:

a) Improves the waveform quality as the level of inverter increases.
b) Reduces the size and rating of filter components.
c) High efficiency due to low switching frequency.
d) Lower dv/dt across switches and generate lower distorted output voltages.
e) Draw input current with very low distortion.
f) Generate smaller common-mode voltage which reduces the stress in the motor bearings.

The several multilevel inverter topologies are: The Neutral point clamped (NPC) inverter, Flying Capacitor Inverter (FCI), and Cascade H-Bridge (CHB) inverter [6]. The NPC inverters are very popular for high voltage and high power applications. Theoretically, NPC topology with any number of levels can be realized. But some of the problems like complexity of switching algorithm, voltage unbalance across capacitors, voltage clamping requirements, and circuit layouts have limits on the level in practical multilevel inverters [7]. In an N-level NPC, each phase leg consists of $2(N-1)$ power switches, $(N-1)\times(N-2)$clamping diodes. The DC bus requires $(N-1)$ bulk capacitors. The line voltage has $(2N-1)$ Levels. At any given time there are (N-1) switches in each leg which are in ON state. Voltage rating of each of the device is assumed to be$\left({V_{dc}}/{(N-1)}\right)$.

In recent years, Fuzzy logic has emerged as an important artificial intelligence tool to characterize and control a system, whose model is not known or ill defined. This paper involves the development of novel methodology to optimize the performance of Mamdani type fuzzy logic controller based on a pre defined objective function. The predefined objective function is optimized by optimizing the normalization parameter, de-normalization parameter and the membership functions of the Fuzzy logic controller.

Recently, there has been a huge interest in the Particle Swarm Optimization (PSO) due to its great potential as an evolutionary algorithm, which is based on the social behaviour of flocks of birds and schools of fish [8].Since it is population based and self adaptive, it has gained an increasing popularity as an efficient alternative to the genetic algorithm (GA) in solving optimization problem. Similar to other population-based optimization method such as the GA, the PSO algorithm starts with random initialization of a population of individuals in the search space. Each particle in the search space is adjusted by its own flying experience and the other particles flying experience to find the global best solution at each generation [8]. This paper proposes an implementation of the PSO method for an off-line tuning of the normalization parameter, de-normalization parameter and membership functions of the fuzzy controller.

## 2. Modeling

### 2.1 Induction Motor

The three-phase squirrel cage induction motor mathematical equations in synchronous rotating reference frames are as follows [1]–[3]:

$$\mathrm{V}_{ds}^{e} = R_s i_{ds}^{e} + p\lambda_{ds}^{e} + \omega_e \lambda_{qs}^{e} \quad (1)$$

$$\mathrm{V}_{qs}^{e} = R_s i_{qs}^{e} + p\lambda_{qs}^{e} + \omega_e \lambda_{ds}^{e} \quad (2)$$

$$0 = R_r i_{dr}^{e} + p\lambda_{dr}^{e} - (\omega_e - \omega_r)\lambda_{qr}^{e} \quad (3)$$

$$0 = R_r i_{qr}^{e} + p\lambda_{qr}^{e} + (w_e - w_r)\lambda_{dr}^{e} \quad (4)$$

Where,

$$\lambda_{ds}^{e} = L_s i_{ds}^{e} + L_m i_{dr}^{e} \quad (5)$$

$$\lambda_{qs}^{e} = L_s i_{qs}^{e} + L_m i_{qr}^{e} \quad (6)$$

$$\lambda_{dr}^{e} = L_r i_{dr}^{e} + L_m i_{ds}^{e} \quad (7)$$

$$\lambda_{qr}^{e} = L_r i_{qr}^{e} + L_m i_{qs}^{e} \quad (8)$$

And electromagnetic torque

$$T_e = \frac{3}{2}\frac{P}{2} L_m (i_{qs}^{e} i_{dr}^{e} - i_{ds}^{e} i_{qr}^{e}) \quad (9)$$

$$w_r = \frac{d\theta_r}{dt} \quad (10)$$

$$T_e = j_m \frac{dw_r}{dt} + B_m w_r + T_l \quad (11)$$

### 2.2 Three-Level NPC Inverter

The three-level NPC inverter with two DC link capacitors $C_1$ and $C_2$ in series and a neutral point O is shown in Fig.1.

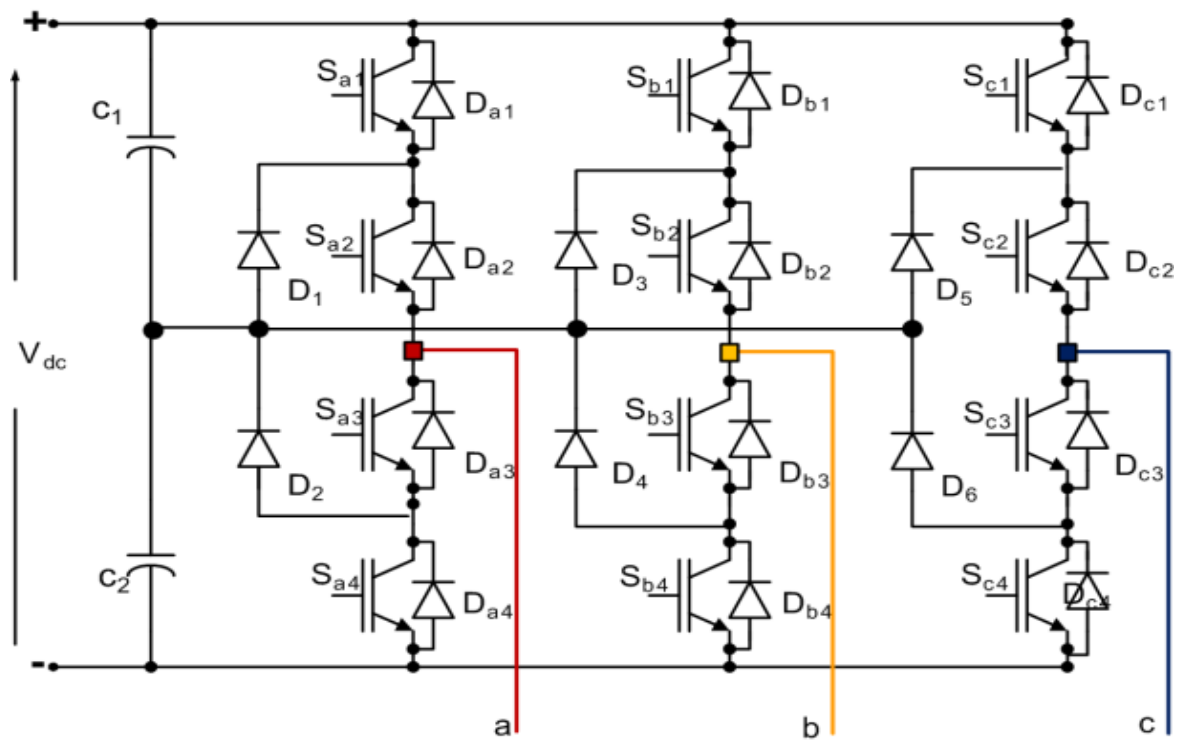

Fig. 1 Three Level NPC Inverter

Table 1: Switching Levels in a Three-Level NPC Inverter

| $S_{i1}$ | $S_{i2}$ | $S_{i3}$ | $S_{i4}$ | $i^{th}$ Pole voltage $V_{io}$ |
|---|---|---|---|---|
| ON | ON | OFF | OFF | $V_{dc}/2$ |
| OFF | ON | ON | OFF | 0 |
| OFF | OFF | ON | ON | $-V_{dc}/2$ |

Each phase of the three-level NPC inverter has two pair of switching devices $S_{i1}$, $S_{i2}$ and $S_{i3}$, $S_{i4}$ in series, where $i = a, b, c$ phases. The center of each pair is clamped to the neutral of the DC link capacitors through the clamping diodes $D_1$, $D_2$, $D_3$, $D_4$, $D_5$ and $D_6$. Table I enumerates the switching states for the semiconductor devices for the $i^{th}$-phase of this inverter. In Table-1, the switching symbols +, 0 and - respectively denote that the $i^{th}$-phase terminal is connected to the positive bus, the neutral point and the negative bus.

### 2.3 Three-Level Hysteresis Current Controller

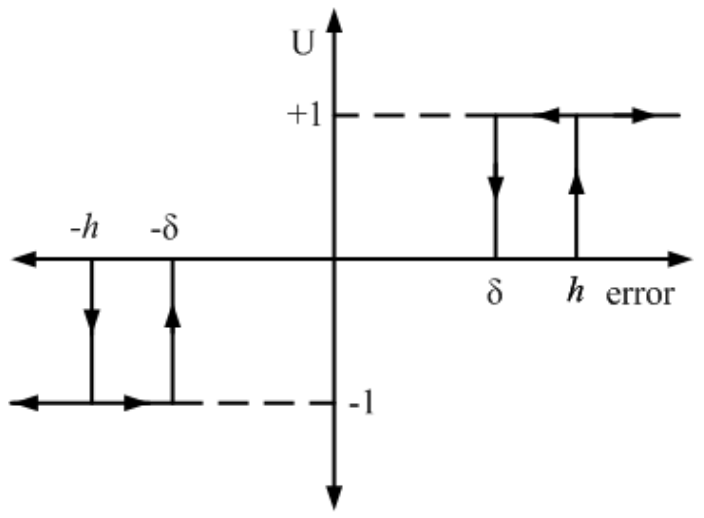

Fig. 2: Three-Level Hysteresis Switching Scheme

An analytical solution of different multilevel PWM techniques for three-level NPC has been presented [9], [10]. Among these techniques, the hysteresis band is used very often because of its simplicity of implementation, fast response current and robust structure [11], [12]. Hysteresis band controller is used to track the line current references. The current errors between the reference and measured currents are used to develop three valid switching states in each inverter leg by the hysteresis band controller.

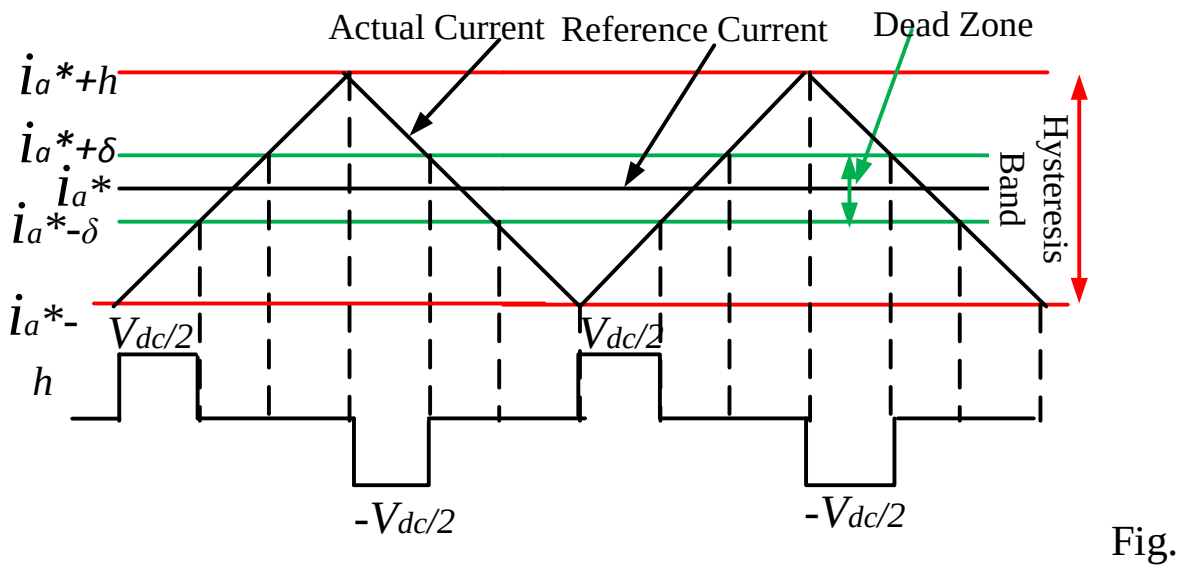

Fig. 3: Three-Level Hysteresis Current Control

To develop a switching scheme for the three-level inverter, the zero voltage level should be applied only at appropriate instants. The switching logic must ensure that there is no successive transition between $\frac{V_{dc}}{2}$ and $-\frac{V_{dc}}{2}$ states, as this will increase the frequency of switching. A dead zone 'δ' is necessary in the hysteresis band 'h', to avoid switching towards two-level scheme, because of finite sampling rate of error. Without the dead zone, when the error becomes zero and is not detected, the opposite polarity of forcing function follows, resulting in a two-level scheme. However, the introduction of dead zone increases the tracking error and has to be chosen to a minimum value, depending on the best sampling speed that can be achieved [13].

If $U$ represents the input state to be applied, $e$ represents error $(i_a^* - i_a)$ and $ce$ represents the change in error the switching logic is governed by equation (12)

If $e > 0$ then

$$U = 1 \;\; for \;\; e \geq h$$
$$U = 0 \;\; for \;\; e \leq \delta$$
$$U = 0 \;\; for \;\; \delta < e < h \; and \; ce > 0$$
$$U = 1 \;\; for \;\; \delta < e < h \; and \; ce < 0$$

Else if $e < 0$ then

$$U = \text{-}1 \;\; for \;\; e \leq \text{-}h$$
$$U = 0 \;\; for \;\; e \geq \text{-}\delta$$
$$U = 0 \;\; for \;\; \text{-}\delta < e < \text{-}h \; and \; ce < 0$$
$$U = \text{-}1 \;\; for \;\; \text{-}\delta < e < \text{-}h \; and \; ce > 0 \qquad (12)$$

The above logic represented in Fig.2 and Fig.3, tracks reference current either in the lower band (through 0 and +1 states) or in the upper band (through 0 and -1 states). Here $U = 1$, means the switch state is $V_{dc}/2$ ; $U = 0$ means the switch state is 0; and $U = -1$, means the switch state is $-V_{dc}/2$ . Similarly the *b*-phase and *c*-phase switching function for the three-phase voltage source inverter can be obtained.

## 2.4 Indirect Vector Control

The indirect vector control is a technique that controls the dynamic speed of Induction motor. Unlike direct vector control, in indirect vector control, the unit vectors are generated in an indirect manner. Fig.4 is the phasor diagram that explains the fundamental principle of indirect vector control. The $d^s - q^s$ axes are fixed on the stator and $d^r - q^r$ axes are fixed on the rotor which rotates at a speed $\omega_r$. Synchronously rotating axes $d^e - q^e$ are rotating ahead of $d^r - q^r$ axes by the positive slip angle $\theta_{sl}$ corresponding to slip frequency $\omega_{sl}$. Thus

$$\theta_e = \int \omega_e dt = \int (\omega_r + \omega_{sl}) dt \qquad (13)$$

For decoupling control $\lambda_{qr} = 0$ or $p\lambda_{qr} = 0$ and $\lambda_r = \lambda_{dr}$ .Substituting the above condition in equations (3), (4), (7) and (8).

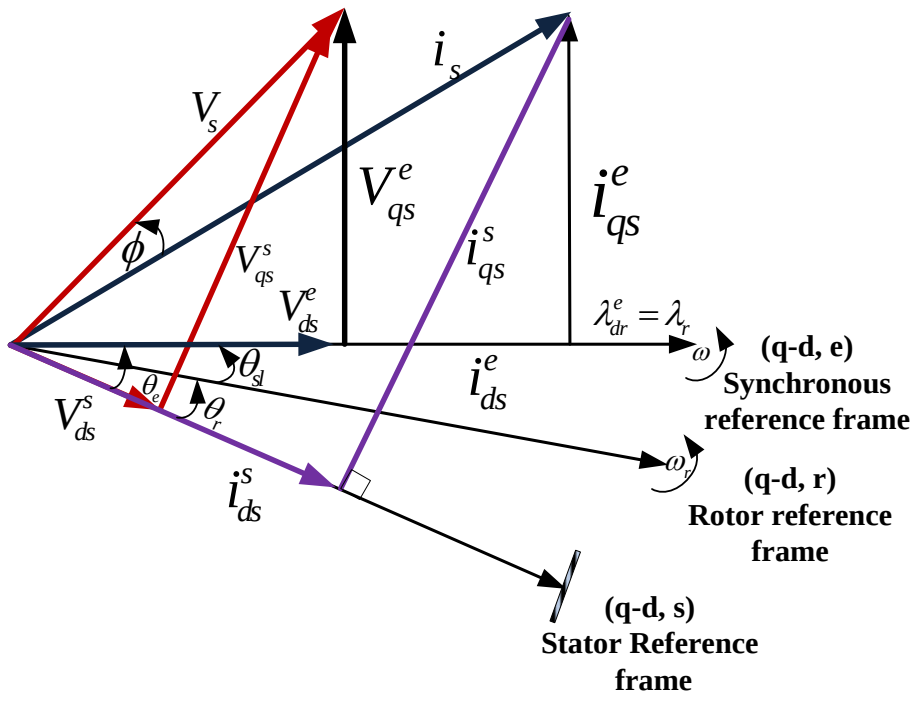

Fig. 4: Phasor diagram of Indirect Vector Control principle

$$\omega_{sl} = \frac{R_r L_m i_{qs}^e}{L_r \lambda_r} \qquad (14)$$

$$T_e = \frac{3}{2}\frac{P}{2}\frac{L_m}{L_r}\lambda_r i_{qs}^e \qquad (15)$$

$$i_{qs}^e = \frac{2}{3}\frac{2}{P}\frac{L_r}{L_m}\frac{T_e}{\lambda_r} \qquad (16)$$

$$i_{ds}^e = \frac{1}{L_m}[\lambda_r + \frac{L_r}{R_r} p\lambda_r] \qquad (17)$$

The equations (14-17) are used to produce an adequate field orientation. These equations could be propagated to the set point variables [14].

$$\omega_{sl}^{*} = \frac{R_r L_m i_{qs}^{e^*}}{L_r \lambda_r^{*}} \tag{18}$$

$$i_{qs}^{e^*} = \frac{2}{3}\frac{2}{P}\frac{L_r}{L_m}\frac{T_e^{*}}{\lambda_r^{*}} \tag{19}$$

$$i_{ds}^{e^*} = \frac{1}{L_m}[\lambda_r^{*} + \frac{L_r}{R_r} p\lambda_r^{*}] \tag{20}$$

If it is accepted that the rotor flux set point is constant then its derivative is zero and the above equation is simplified as

$$i_{ds}^{e^*} = \frac{\lambda_r^{*}}{L_m} \tag{21}$$

Using the above equations the block diagram of indirect vector control of induction motor drive is as shown in Fig.5.

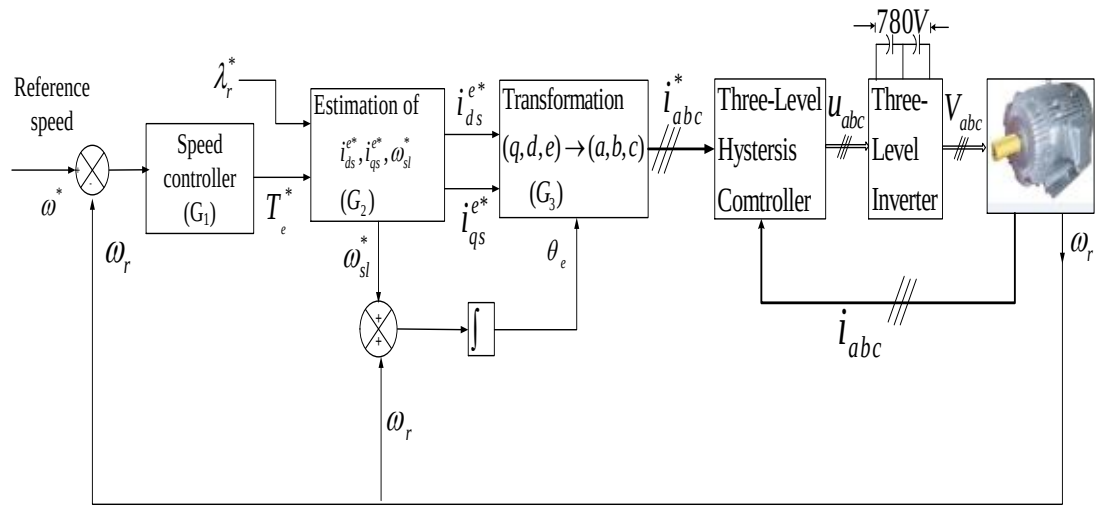

Fig. 5: Block diagram of Indirect Vector Control of IM

It contains three principal blocks, $G_1$ used as speed controller, $G_2$ used for estimation of $i_{ds}^{e^*}$, $i_{qs}^{e^*}$, $\omega_{sl}^{*}$ and $G_3$ used for current co-ordinate transformation (q, d, e) to (a, b, c).

## 3. Fuzzy Logic Speed Controller

The speed controller block G1 is proposed to be a Mamdani type fuzzy controller having five blocks namely normalizer, fuzzifier, inference mechanism, de-fuzzifier, and de-normalizer as shown in Fig.6. [15].

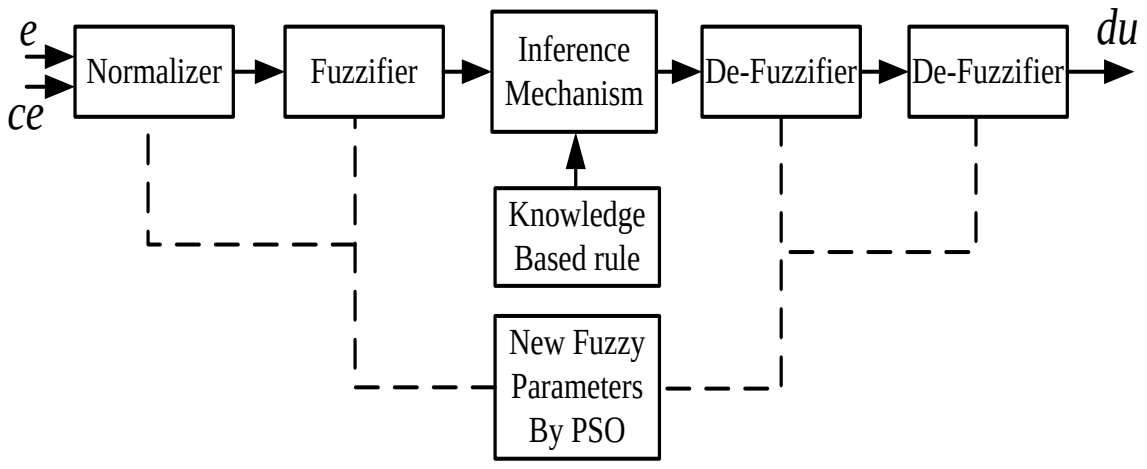

Fig. 6: Block Diagram of Fuzzy Controller

### 3.1 Normalizer and De- normalizer

In closed loop control system the use of error *(e)* and the change in error *(ce)* as controller input is a universal approach. Therefore the fuzzy controller has two inputs, error and change in error (e, ce) and one output (du) as shown in Fig. The error is the difference between the reference speed and the actual rotor speed.

$$Input_1 = e(t) = \omega^{*}(t) - \omega(t) \tag{22}$$

$$Input_2 = ce(t) = e(t) - e(t-1) \tag{23}$$

Where, $\omega$ is the actual speed and $\omega^{*}$ is the reference speed. Two normalization parameters *($k_1$, $k_2$)* for inputs *(e, ce)* and one de-normalization parameter *($k_3$)* for output (*du*) are defined. In normalization process the input variables are scaled in the range of *(-1, +1)* and in de-normalization process the output values of fuzzy controller are converted to a value depending on the terminal control element. The determination of normalization and de-normalization parameters of fuzzy controller is important for system stability.

### 3.2Fuzzifier and De-fuzzifier

The fuzzifier processes the crisp input values *(e, ce)* and convert them into fuzzy values. Also the fuzzy values obtained in fuzzy inference mechanism are converted to crisp output *(du)* value by a de-fuzzifier. Here, a triangular fuzzy membership function is defined for each input and output values by seven clusters. For seven clusters in the membership functions, seven linguistic variables are defined as: Negative Big (NB), Negative Medium (NM), Negative Small (NS), Zero (Z), Positive Small (PS), Positive Medium (PM), Positive Big (PB). Fig.11 shows the membership functions used to fuzzify two input values *(e, ce)* and de-fuzzify output *(du)* of the realized fuzzy controller. The peak or bottom points of the membership functions to be tuned are $a_1$ and $a_2$ for error *(e)*, $b_1$ and $b_2$ for change in error *(ce)* and $c_1$ and $c_2$ for output *(du)*. Therefore the design of fuzzy controller requires the optimization of nine parameters *($k_1$, $k_2$, $k_3$, $a_1$, $a_2$, $b_1$, $b_2$, $c_1$, $c_2$)*.

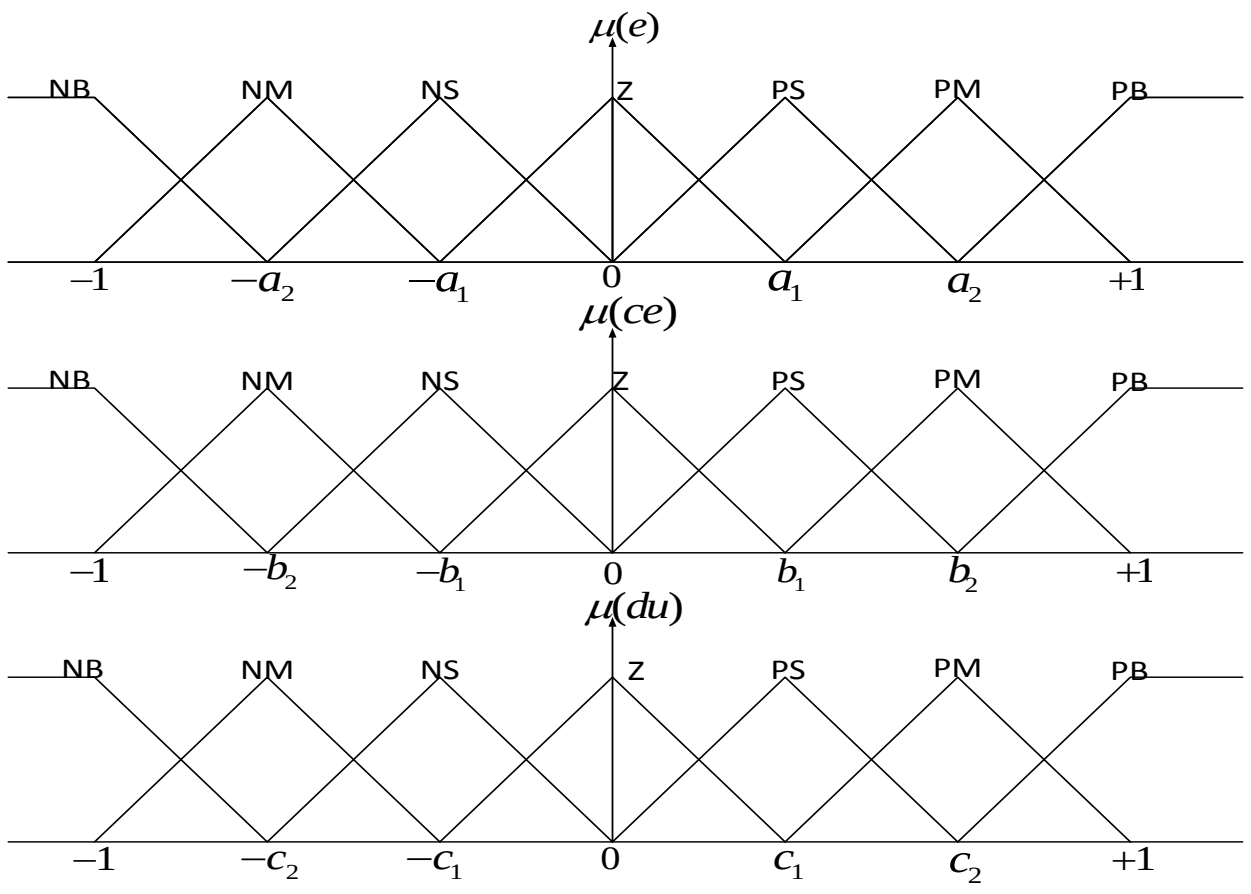

Fig. 7: Membership functions of Inputs and output

In this work the centre of gravity or centroid method is used for de-fuzzification. As a result the control increment is obtained by the equation [16].

$$du = \frac{\sum_{i=1}^{m} d_i A(\mu_i)}{\sum_{i=1}^{m} A(\mu_i)} \tag{24}$$

Here $d_i$ is the distance between $i^{th}$ fuzzy set and the centre, $A(\mu_i)$ is area value of $i^{th}$ fuzzy set.

## 3.3 Knowledge Base and Inference Mechanism

The rule definition is subjective and based on expert's knowledge and experiences. It establishes the relationship between outputs with inputs [17]. For the system with two inputs and seven membership functions in each leads to forty nine combination of these inputs, in which there are forty nine rules. The rules are like:

*R1. If e = NB and ce = NB Then du is NB*
*or*
*R2. If e = NB and ce = NM Then du is NB*
*or......*
*R49. If e = PB and ce = PB Then du is PB*

The rules are represented by a matrix called matrix inference shown in Table 2. A feature of the rule base used is the symmetry across the diagonal. This feature occurs in systems where the physical behaviors of the system exhibit symmetry, which is consistent in case of speed control of Induction, motor.

The developed fuzzy logic uses the inference method for each rule given by the relation

$$\mu_i(du) = \min(\mu_i(e), \mu_i(ce)); \quad i = 1, 2, ......49 \tag{25}$$

Table 2: Fuzzy Linguistic Rule Table

| | | **Error (e)** | | | | | | |
|---|---|---|---|---|---|---|---|---|
| | | **NB** | **NM** | **NS** | **Z** | **PS** | **PM** | **PB** |
| **Change in Error (ce)** | **NB** | NB | NB | NB | NB | NM | NS | Z |
| | **NM** | NB | NB | NB | NM | NS | Z | PS |
| | **NS** | NB | NB | NM | NS | Z | PS | PM |
| | **Z** | NB | NM | NS | Z | PS | PM | PB |
| | **PS** | NM | NS | Z | PS | PM | PB | PB |
| | **PM** | NS | Z | PS | PM | PB | PB | PB |
| | **PB** | Z | PS | PM | PB | PB | PB | PB |

Therefore, the resulting membership function is given by

$$\mu(du) = \max(\mu_1(du), \mu_2(du), .......\mu_{49}(du)) \tag{26}$$

# 4. Fuzzy Logic Speed Controller Based on Particle Swarm Optimization

## 4.1 Particle Swarm Optimization

Particle Swarm Optimization is a population based stochastic optimization technique, inspired by social behaviour of bird flocking or fish schooling. In PSO system the individuals called particles, fly around in a multidimensional search space and change their position with time. During its flight, each particle adjusts its position according to its own experience and according to the experience of neighbouring particle. The position or value corresponding to its own experiences called $\boldsymbol{P_{best}}$ and corresponding to the experience of neighbouring particle is called $\boldsymbol{G_{best}}$. The search for the optimal position advances as the particles' velocities and positions are updated. The fitness of each particle's position and iteration is calculated using a pre defined objective (fitness) function and the velocity of each particle is updated using the $\boldsymbol{P_{best}}$ and $\boldsymbol{G_{best}}$, which were previously defined. The velocity of $\boldsymbol{i^{th}}$ particle can be modified by the following equation.

$$v_i(n+1) = \omega(n) \times v_i(n) + C_1 \times r_1(n) \times \left(P_{best_i} - x_i(n)\right) + C_2 \times r_2(n) \times \left(G_{best_i} - x_i(n)\right) \tag{27}$$

Where, $v_i(n)$ is the velocity of $\boldsymbol{i^{th}}$ particle at iteration *n*, $v_i(n+1)$ is the velocity of $\boldsymbol{i^{th}}$ particle at iteration *(n + 1)*, $r_1(n)$ and $r_2(n)$ are random numbers with uniform distribution in the interval [0, 1], $\omega(n)$ is the momentum or inertial weight constant given by [18]

$$\omega(n) = \omega_{max} - \left\{\frac{\omega_{max} - \omega_{min}}{n_{max}}\right\} \times n \tag{28}$$

Here $n_{max}$ is the maximum number of iteration, $\omega_{max}$ and $\omega_{min}$ are the maximum and minimum weights respectively. Appropriate values of $\omega_{max}$ and $\omega_{min}$ are 0.9 and 0.4 respectively [19]. The values $C_1$ and $C_2$ are two positive constants represent the social and cognitive accelerations for the $\mathbf{P_{best}}$ and $\mathbf{G_{best}}$ positions, respectively. Varying these parameters has the effect of varying the strength of the pull towards the two bests. Values of $C_1 = C_2 = 0$ mean that both the cognitive and social accelerations are absent, and particles keep moving at their current speed until they hit a boundary of the search space (assuming no inertia) [20].With $C_1 > 0$ and $C_2 = 0$, each particle searches for the best position in its neighbourhood, and replaces the current best position if the new position is better [20]. However, with $C_2 > 0$ and $C_1 = 0$, the entire swarm is attracted to a single point, $\mathbf{G_{best}}$. Furthermore, having $C_1 >> C_2$ causes each particle to be attracted to its own personal best position to a very high extent, resulting in excessive wandering. On the other hand, $C_2 >> C_1$ results in particles being more strongly attracted to the global best position, thus causing particles to rush prematurely towards optima [20]. It is demonstrated that the particle swarm is only stable and guaranteed to converge to a stable equilibrium point if the following conditions are satisfied [21].

$$0 < (C_1 + C_2) < 4 \tag{29}$$

$$\frac{(C_1 + C_2)}{2} - 1 < \omega(n) < 1 \tag{30}$$

However, whether or not this point is actually the global minimum cannot be guaranteed, and its acceptability as a solution should be verified. The position of $i^{th}$ particle at iteration $n$ is $x_i(n)$. The modified position at iteration $(n+1)$ is given by

$$x_i(n+1) = x_i(n) + v_i(n+1) \tag{31}$$

### 4.2 Optimization of Fuzzy Controller

The Particle Swarm Optimization is applied to automate and optimize the fuzzy controller design process. The normalization parameters $(k_1, k_2, k_3)$ and the parameters of the membership functions $(a_1, a_2, b_1, b_2, c_1, c_2)$ are optimized by optimizing a properly defined objective or fitness function [22], [23]. In the context of optimization our goal is to have a speed response with a short rise time, small overshoot and near zero steady state error. In this respect a multiple objective function is defined as

$$F = \int_0^t |e|\,dt + \int_0^t |e|\,t\,dt \tag{32}$$

Where, the first term is the measure of fast dynamic response and the second term is the measure of steady state error.

Thus the purpose of PSO algorithm is to minimize the objective function. The PSO based approach to find the minimum value of objective function is as shown in Fig.8. The input parameters of the proposed PSO algorithm are: $n_{\max} = 100, n_{pop} = 30, n_{\mathrm{var}\,s} = 9, C_1 = 0.5, C_2 = 1.25,$ $\omega_{\max} = 0.9, \omega_{\min} = 0.4, 0 \le (a_1, a_2, b_1, b_2, c_1, c_2) \le 1,$ $a_2 > a_1, b_2 > b_1, c_2 > c_1, 0 \le k_1 \le 6.67e-3,$ $0 \le k_2 \le 1, 0 \le k_3 \le 6,$ Stall generation =20 and Function tolerance $=1e-6$.

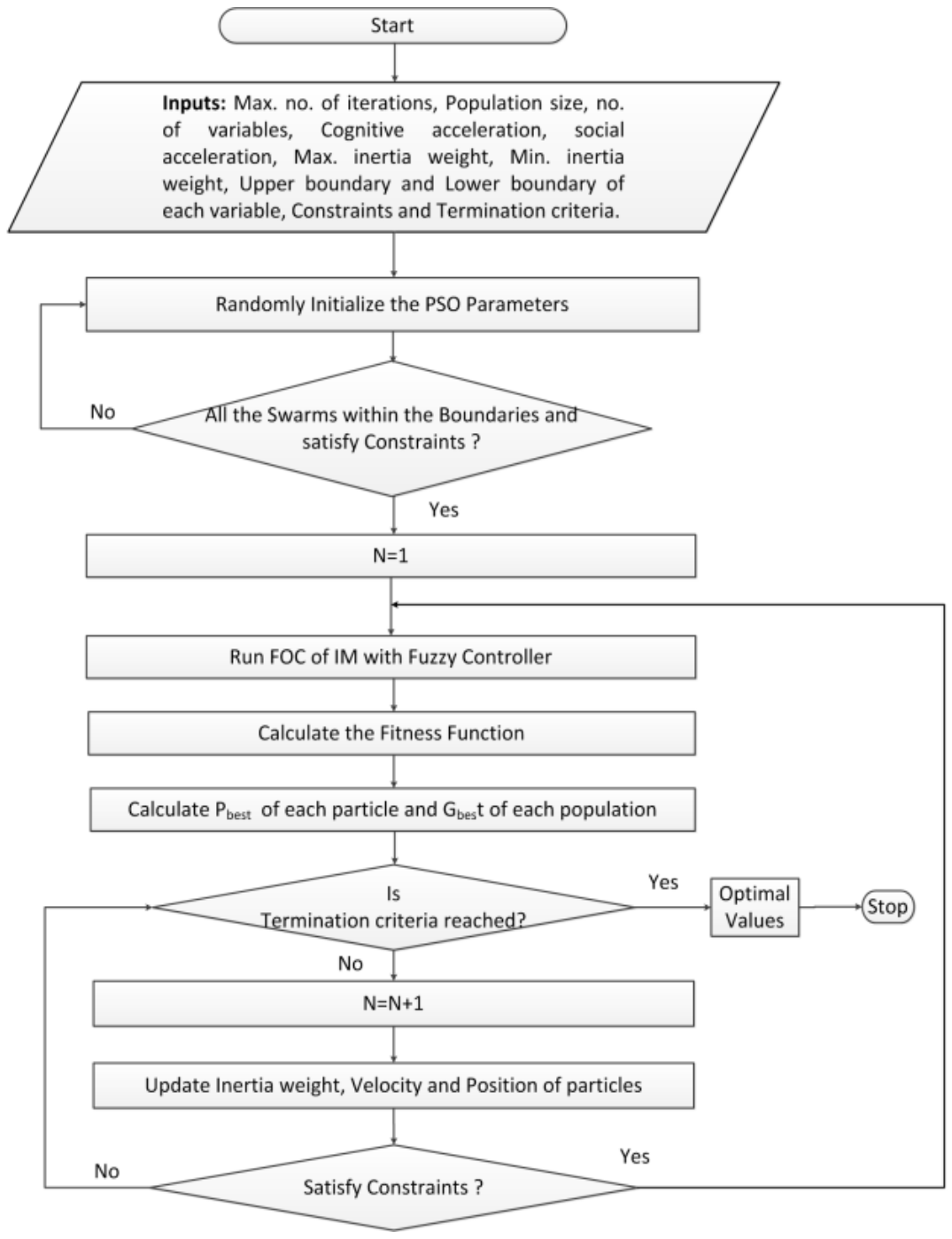

Fig. 8: Flowchart of the PSO algorithm

## 5. Simulation Results and Discussion

Complete simulation model for vector controlled Induction motor (IM) drive of the proposed scheme is developed using MATLAB/ SIMULINK. The motor parameters are:

Rated Power $P_{rated} = 50HP$, Rated Voltage $V = 480volt$, Rated Frequency $F = 50Hz$, Pair of poles $P=2$, Stator Resistance $R_s = 0.087\Omega$, Rotor Resistance $R_r = 0.228\Omega$, Stator Inductance $L_s = 0.8mH$, Rotor Inductance $L_r = 0.8mH$, Mutual Inductance $L_m = 34.7mH$, Moment of Inertia $J = 1.662Kg.m^2$.

Fig.9 shows the scores of the fitness function corresponding to different generation in PSO. The PSO is terminated at 41 generations as the termination criteria reached. The termination criteria of the algorithm is either the maximum generations reached or the weighted average change in the fitness function value over Stall generations is less than function tolerance. The values of nine parameters used in fuzzy controller and their optimized values by particle swarm optimization are shown in Fig.10. The particle swarm optimized input and output membership functions are in Fig.11.

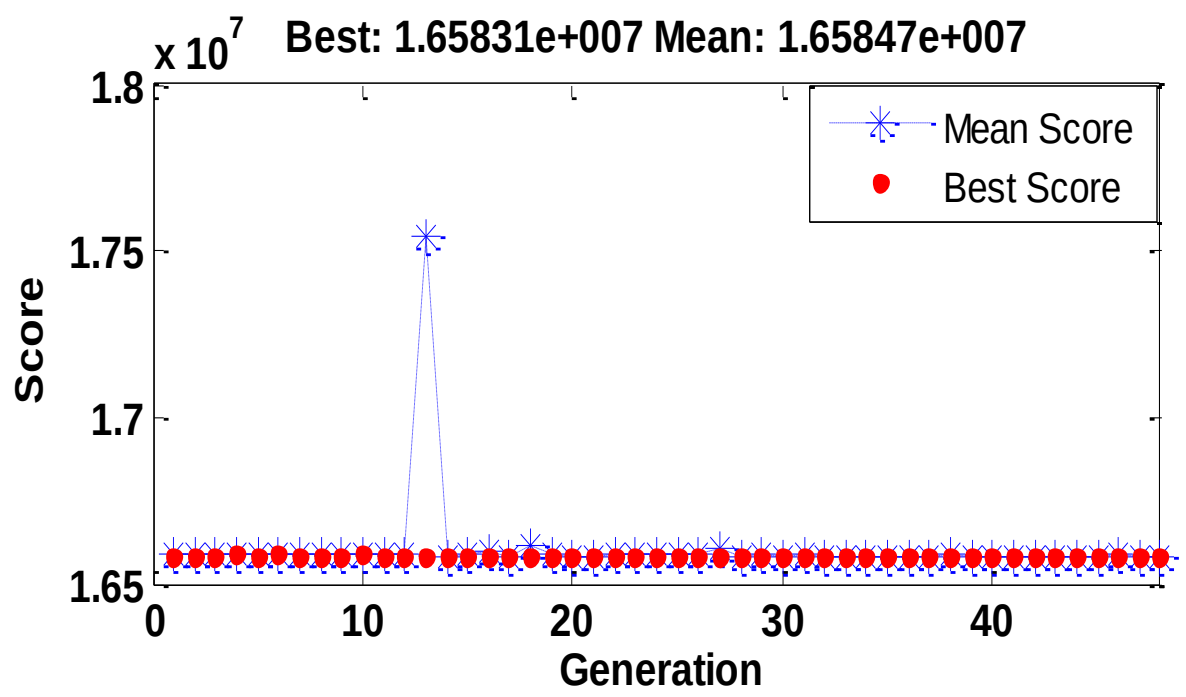

Fig. 9: Fitness Score versus Generation

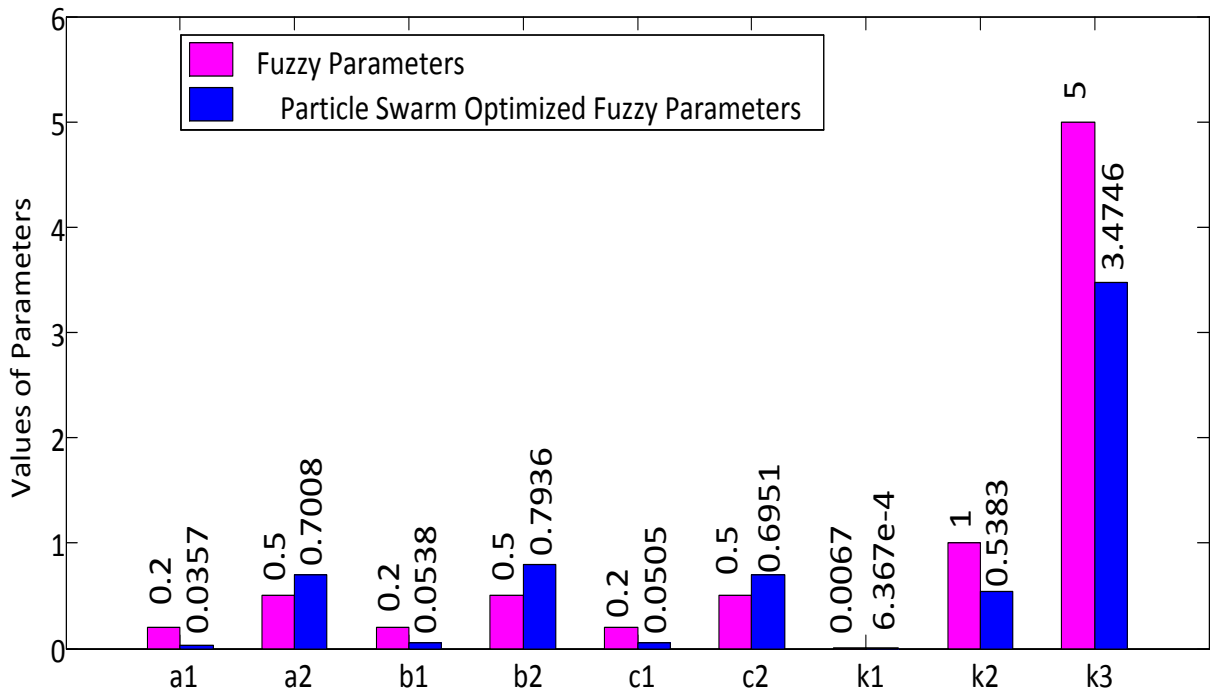

Fig. 10: Conventional and PS Optimized fuzzy Controller parameters

Fig.12 shows the pole voltage, line voltage and the line currents (stator line currents) of the three-level inverter under steady state condition. The line currents are sinusoidal with almost negligible ripple. Fig.13 shows the speed tracking performance of the motor following a trapezoidal speed reference. The speed tracking experiment is on no load condition. The motor speed almost tracks the reference speed in both the direction. The ripple content in the torque during the transition is comparatively reduced with PSO fuzzy controller as compared to simple fuzzy controller as shown in Fig.14

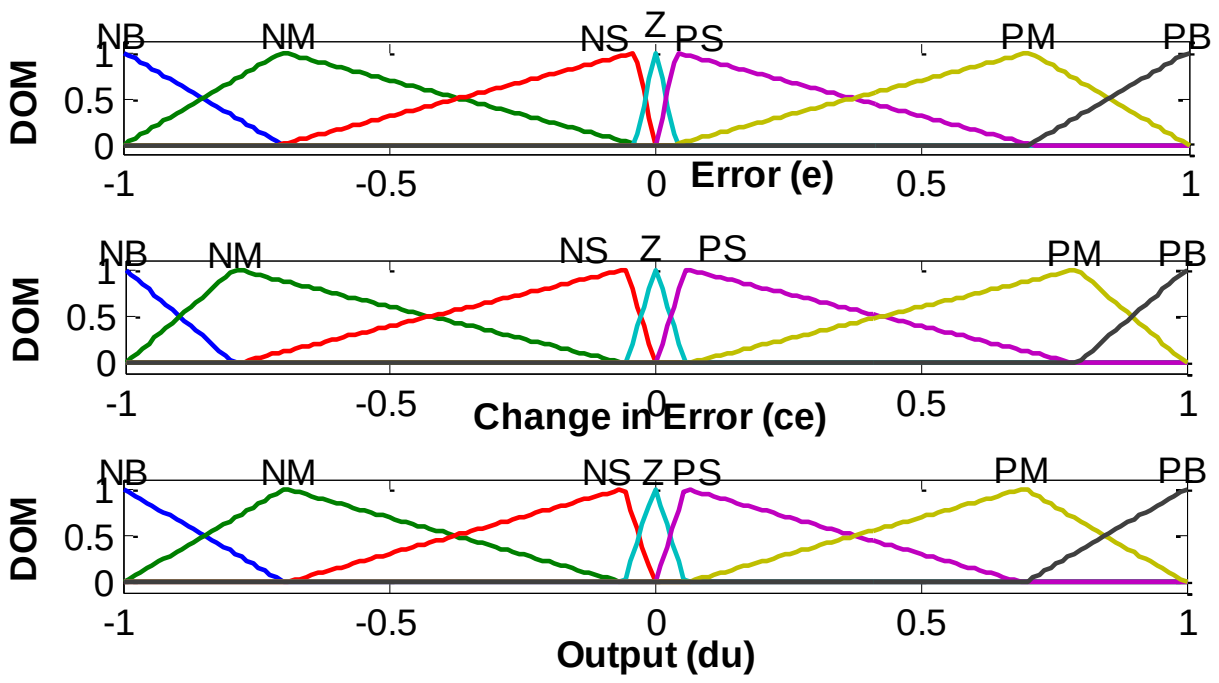

Fig. 11: PS Optimized Input Output Membership Functions of fuzzy controller

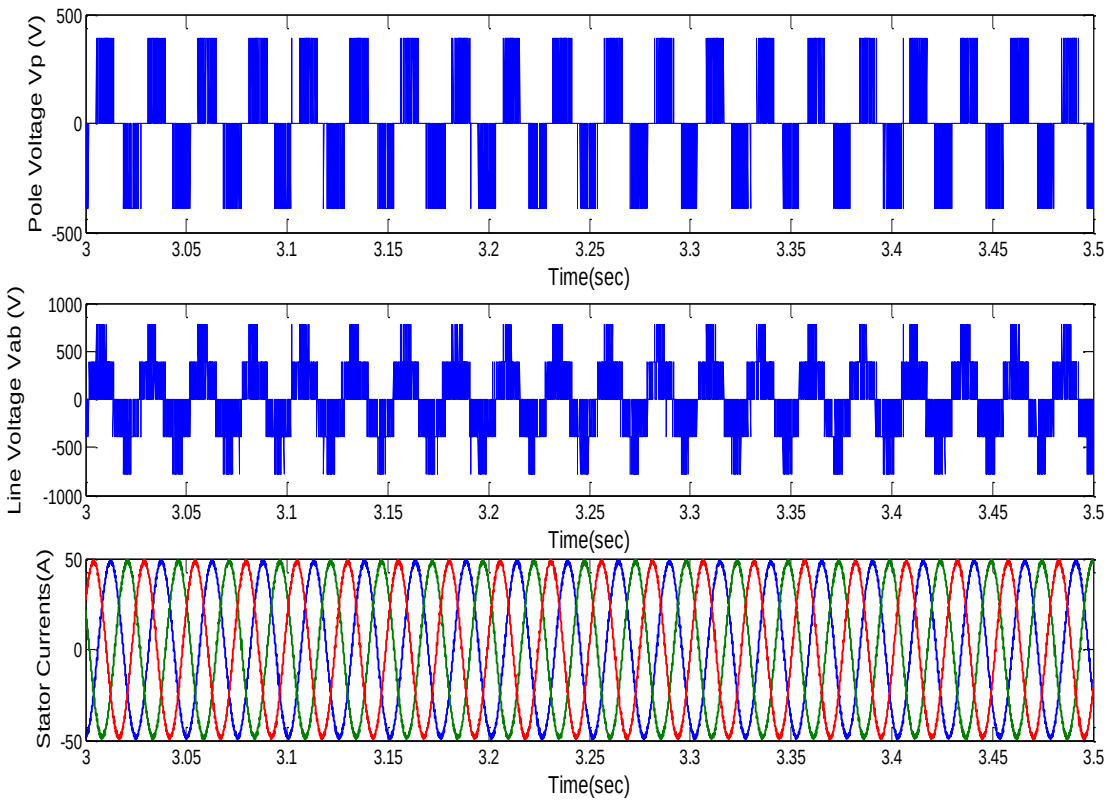

Fig. 12: Inverter Voltages and Currents

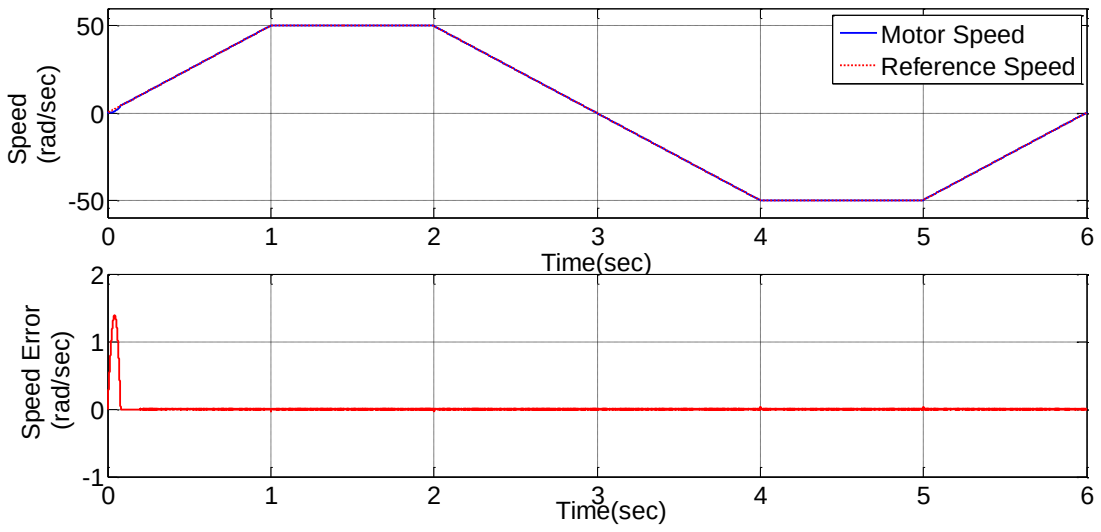

Fig. 13: Trapezoidal Speed Tracking

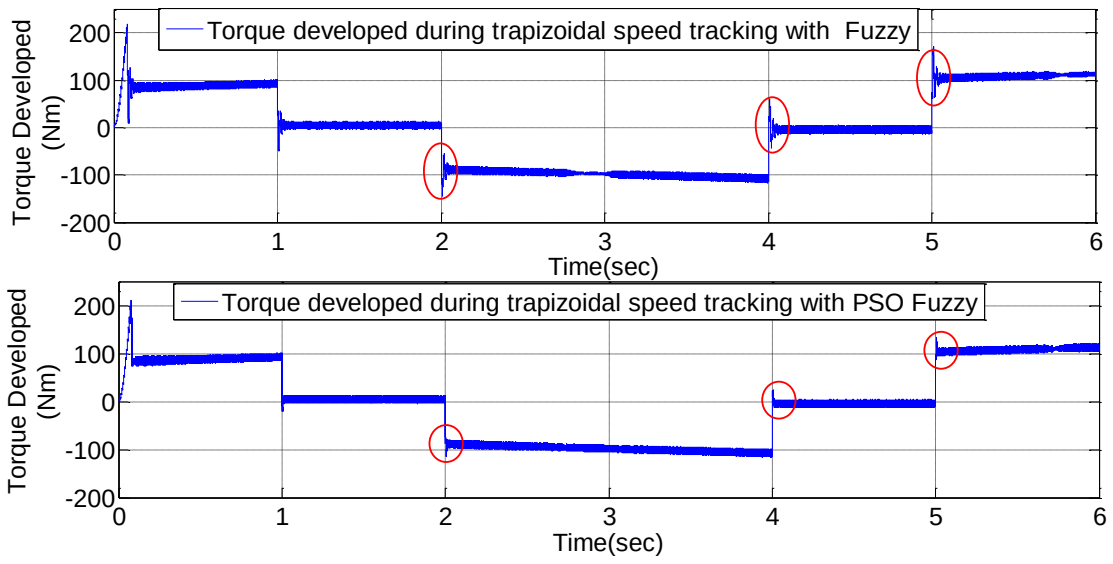

Fig. 14: Torque Developed in Trapezoidal Speed Tracking

Fig. 15 and Fig.16 show the performance of motor for the constant reference speed of $120\, rad/sec$ with constant load torque $100\ Nm$ in both fuzzy and PSO fuzzy speed controller. The ripple in speed and torque, when the motor achieves the reference speed is nearly zero in case of PSO fuzzy as compared to fuzzy controller. Fig.17 and Fig.18 show the performance of motor when the load torque is suddenly changed from $50\ Nm$ to $150 Nm$ and then from $150 Nm$ to $80\ Nm$ at constant reference speed $100\ rad/sec$.

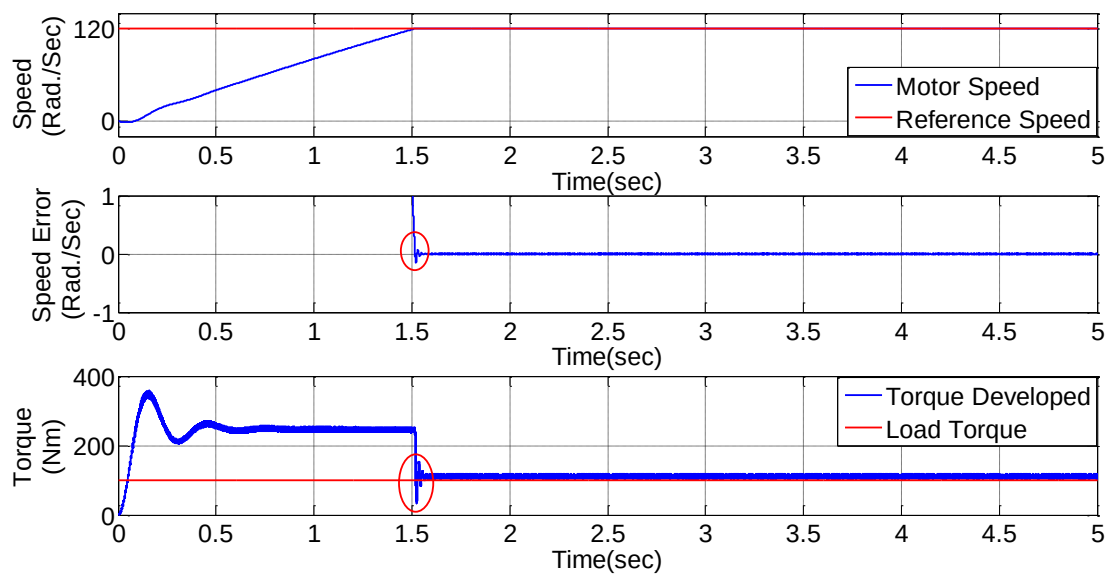

Fig. 15: Performance under constant speed and constant torque with fuzzy

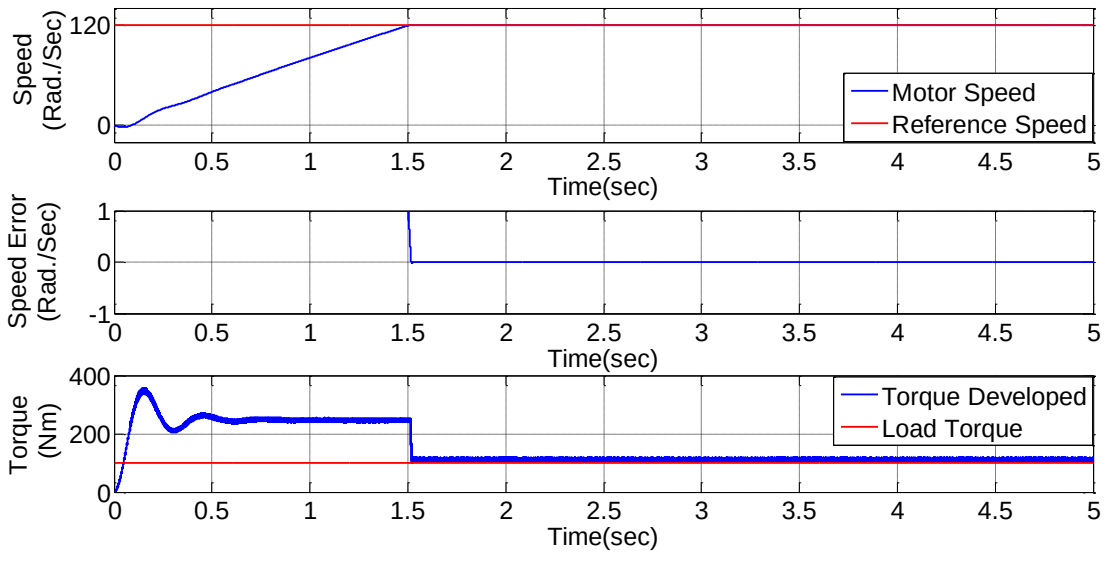

Fig. 16: Performance under constant speed and constant torque with PSO fuzzy

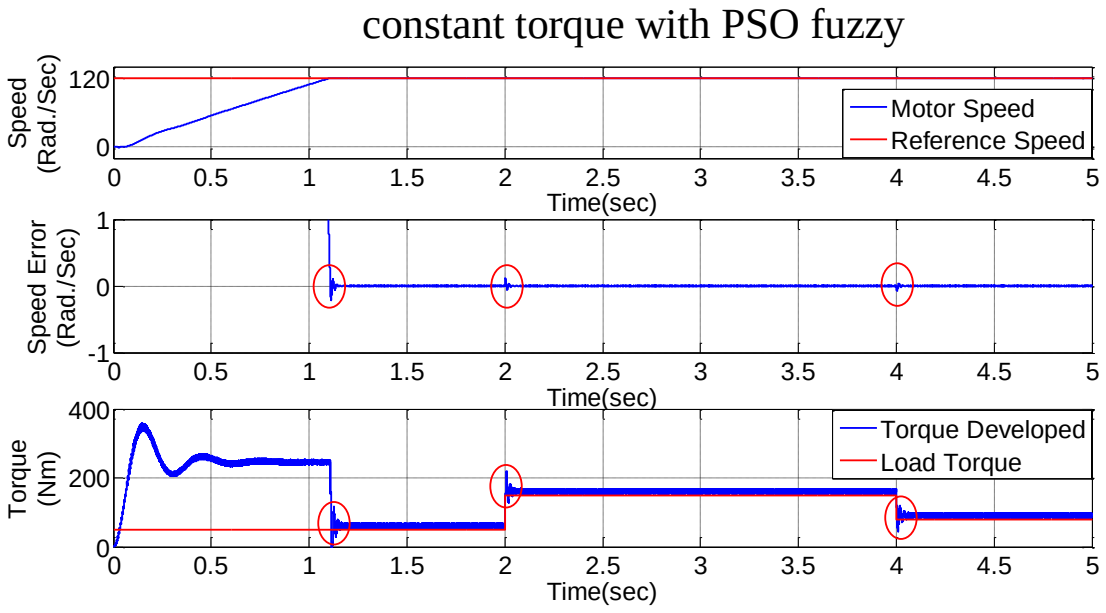

Fig. 17: Performance under constant speed and variable torque with fuzzy

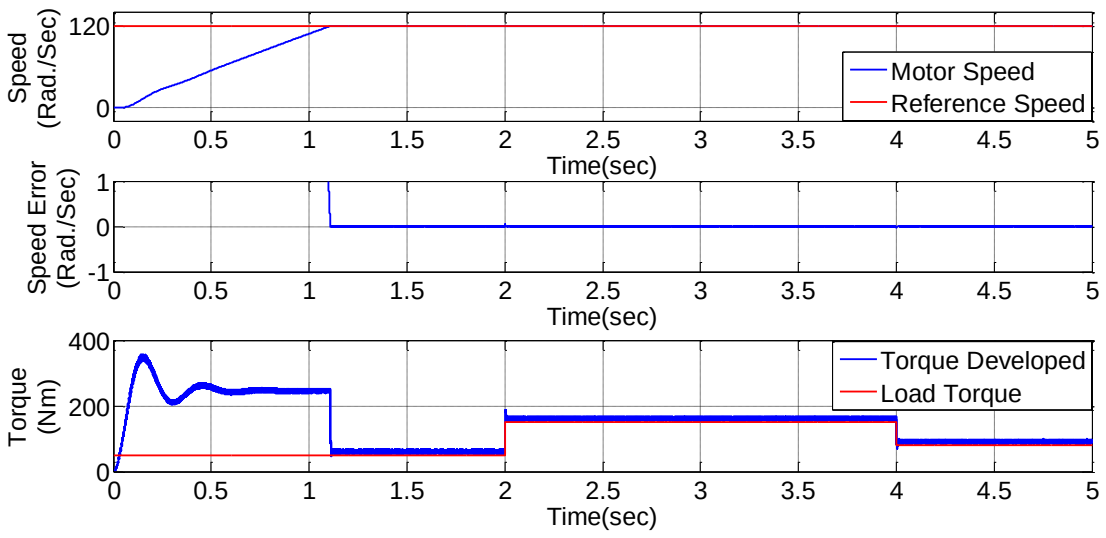

Fig. 18: Performance under constant speed and variable torque with PSO fuzzy

Fig.19 and Fig.20 show the performance of the motor when the reference speed is suddenly changed from $50 rad/sec$ to $120\ rad/sec$ and then from $120\ rad/sec$ to $80\ rad/sec$ with a constant load torque of $100\ Nm$. Fig.21 and Fig.22 show the performance of the motor with variable speed and variable torque. The speed is increased from $50\ rad/sec$ to $120\ rad/sec$ and then decreased from $120\ rad/sec$ to $80\ rad/sec$ with a variable load torque. The load torque is increased from $30 Nm$ to $150\ Nm$ and then it is decreased to $100\ Nm$. In all these cases the performances are better with PSO fuzzy as compared to fuzzy controller.

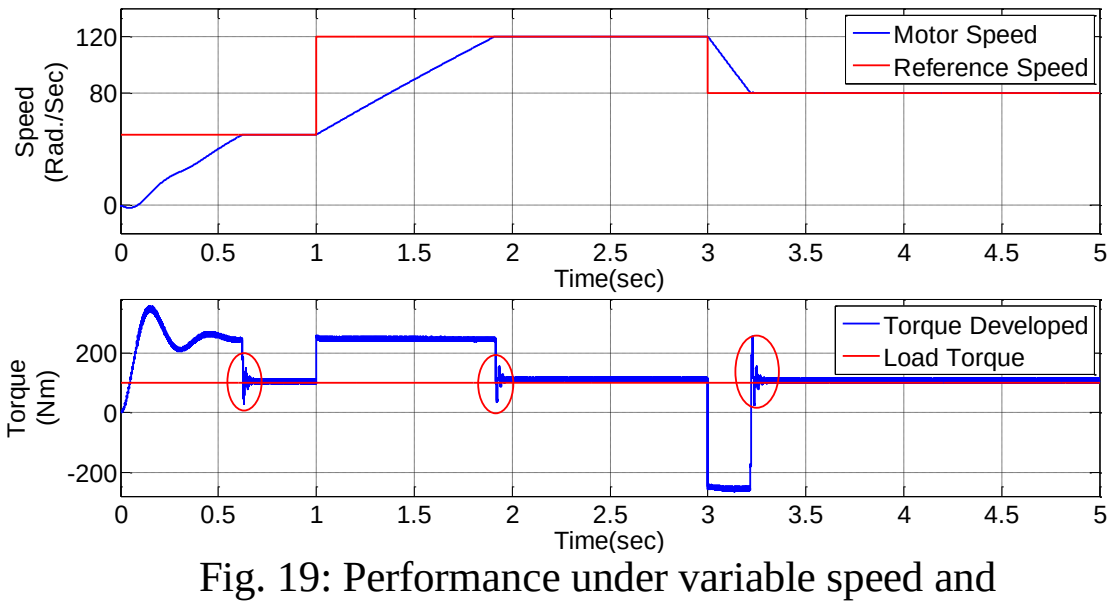

Fig. 19: Performance under variable speed and constant torque with fuzzy

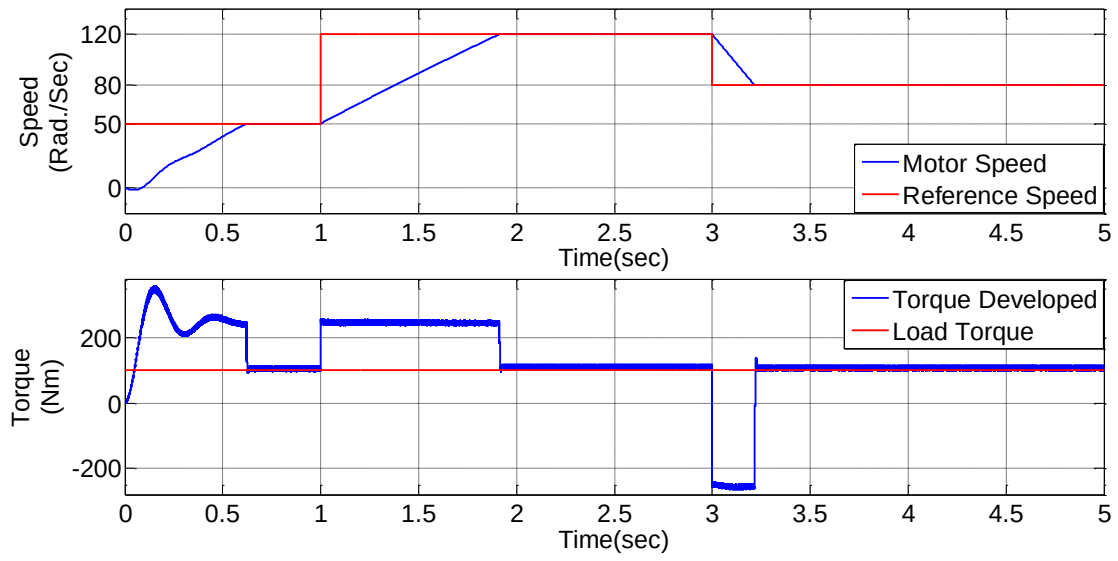

Fig. 20: Performance under variable speed and constant torque with PSO fuzzy

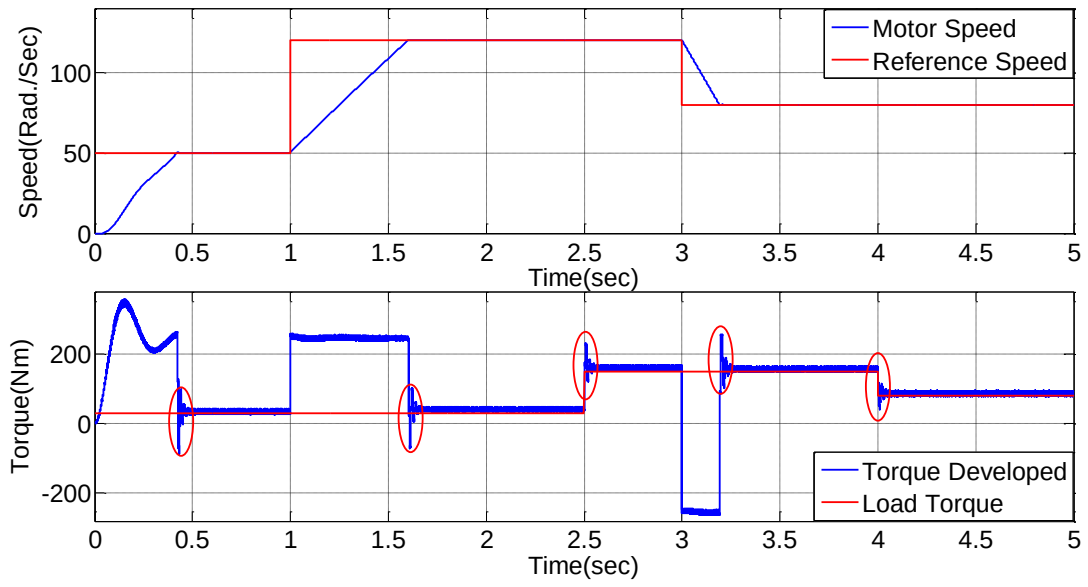

Fig. 21: Performance under variable speed and variable torque with fuzzy

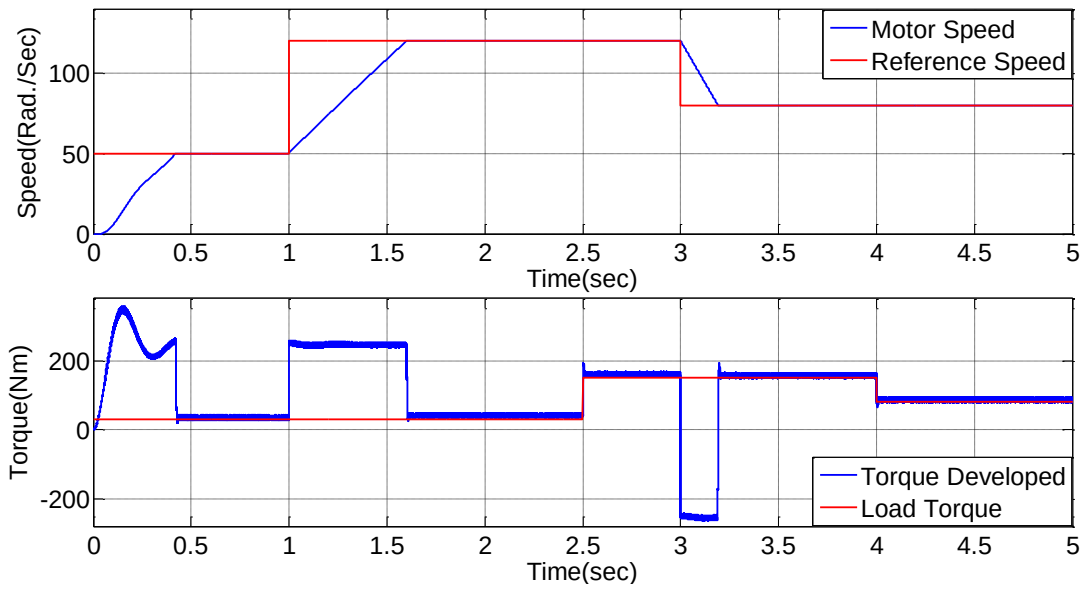

Fig. 22: Performance under variable speed and variable torque with PSO fuzzy

## 6. Conclusions

The optimal fuzzy controller has been designed off-line using techniques of Particle Swarm Optimization for indirect vector control of multilevel inverter fed Induction motor. It achieves good pursuit of reference speed, starting without overshoot and rapid rejection of disturbances with a low drop-out speed. By comparison with fuzzy logic controller, it testifies that this method is not only robust, but also can improve dynamic performance of the system.

## References

[1] B. K. Bose, Modern Power Electronics and AC Drives, Inc. Upper Saddle River, NJ-07458: Prentice-Hill PTR Companies, 2002.

[2] P. Vas, Vector control of AC machines, New York: Clarendon, 1990.

[3] R. Krishnan, Electric motor drives modeling, analysis and control. New Delhi: PHI Pvt. Ltd, 2003.

[4] A. Nabae, I. Takahasi, and H. Akagi, "A neutral point clamped PWM inverter", IEEE Transaction on Industry Applications, Vol. 1A, No. 17,Sept./Oct 1981, pp. 518–523.

[5] J. Rodrguez, J.-S. Lai, and F. Z. Peng, "Multilevel inverters: A survey of topologies, controls, and applications", IEEE Transaction on Industrial Electronics, Vol. 49, No. 4, Aug. 2002.

[6] S. Lai and F. Z. Peng, "Multilevel converters- a new breed of power converters", IEEE Transaction on Industry Applications, Vol. 32, No. 3, May 1996, pp. 509–517.

[7] X. Yuan and I. Barbi, "A new diode clamping multilevel inverter," IEEE, conference, 1999, pp. 495–501.

[8] J. Kennedy and R. C. Eberhart, "Particle swarm optimization," Proceedings of the IEEE Int. Conf. on Neural Networks, Perth, Australia,1995, pp. 1942–1948.

[9] M. Kazmierkowski and L. Malesani, "Current control techniques for three-phase voltage-source pwm converters: A survey," IEEE Transaction Ind. Electron, Vol. 45, No. 5, 1998, pp. 691–703.

[10] A. Shukla, A. Ghosh, and A. Joshi, "Hysteresis modulation of multilevel inverters," IEEE Transaction on Power Electronics, Vol. 26, No. 5, May 2011, pp. 1396–1408.

[11] J. Zeng, C. Yu., Q. Qi., Z. Yan., Y. Ni., B. Zhang, S. Chen, and F. F. Wu., "A novel hysteresis current control for active power filter with constant frequency", Electric Power Systems Research, Vol. 68, 2004, pp. 75–82.

[12] M. Milosevic, "Hysteresis current control in three-phase voltage source inverter",Zurich: Technical Report, 2003.

[13] S. Srikanthan, M. Mishra, and R. Rao, "Improved hysteresis current control of three-level inverter for distribution static compensator application", IET Power Electronics, Vol. 2, No. 5, 2009, pp. 517–526.

[14] S. K. Sahu, D. D. Neema, and T. V. Dixit, "Indirect vector control of induction motor using ANN estimator and ANFIS controller," International Journal of Computer Applications, Vol. 66, No. 14, March 2013.

[15] M. N. Uddin, T. S. Radwan, and M. A. Rahman, "Performances of fuzzy-logic-based indirect vector control for induction motor drive," IEEE Transaction on Industry Applications, Vol. 38, No. 5, Sept./Oct 2002, pp. 1219–1225.

[16] C. C. Lee, Fuzzy Logic in Control Systems: Fuzzy Logic controller Part 1 and Part 2. New Delhi: IEEE Press, 1990.

[17] Z. Zhao, M. Tomizuka, and S. Isaka, "Fuzzy gain scheduling of PID controllers," IEEE Transactions on Systems, Man and Cybernetics, Vol. 23, No. 5, Sept./Oct 1993, pp. 1392–1398.

[18] Y. Bekakra and D. B. Attous, "Sensorless speed based on MARS with tuning of PI speed controller in FOC of induction motor drive using PSO," World Academy of Science, Engineering and Technology, Vol. 60, 2011, pp. 1550–1555.

[19] R. Eberhart. and Y. Shi, "Comparing inertial weights and constriction factor in particle swarm optimization," Proceeding of the International Congress on Evaluationing Computation, 2000, pp. 84–88.

[20] E. AP, Fundamentals of Computational Swarm Intelligence. John Wiley Sons, 2005.

[21] R. Perez and K. Behdinan, "Particle swarm approach for structural design optimization", Computers and Structures, Vol. 85, 2007, pp. 1579–88.

[22] V. Donescu, D. Neacsu, G. Griva, and F. Profumo, "A systematic design method for fuzzy controller for brushless dc motor drives," Proc. of the 27th. IEEE Annual Power Electronics Specialists Conference, 1996, pp. 689– 694.

[23] F. D. S. Cardoso, J. F. Martins, and V. F. Pires, "A comparative study of a PI, neural network and fuzzy genetic approach controllers for an ac drive," IEEE 5$^{th}$International Workshop on Advanced Motion Control AMC, Coimbra, 1998, pp. 375–380.